\documentclass[10pt,twocolumn,letterpaper]{article}

\usepackage[pagenumbers]{iccv} 

\usepackage[dvipsnames]{xcolor}
\usepackage{graphics} 
\usepackage{epsfig} 
\usepackage{times} 
\usepackage{amsmath} 
\usepackage{amssymb}  
\usepackage{enumitem}
\usepackage{mathtools}
\usepackage{algorithm}
\usepackage{algorithmic}
\usepackage{wrapfig}
\usepackage{colortbl}
\usepackage{listings}
\usepackage{afterpage}
\usepackage{relsize}
\usepackage{multicol}
\usepackage{multirow}
\usepackage{wasysym}
\usepackage{booktabs}
\usepackage{caption}
\usepackage{subcaption}
\usepackage{sidecap}
\usepackage{pifont}
\usepackage[nodisplayskipstretch]{setspace}
\usepackage{xcolor}
\usepackage{url}
\usepackage{arydshln}

\newcommand\mypar[1]{\par\vspace{-0.2mm}\noindent\textbf{#1}\;\;}
\definecolor{LightGrey}{rgb}{0.92,0.92,0.92}
\definecolor{Myred}{rgb}{1.00,0.12,0.36}
\definecolor{Myblue}{rgb}{0,0.60,0.87}
\newcommand{\ourwork}{PaintScene4D\xspace}

\definecolor{iccvblue}{rgb}{0.21,0.49,0.74}
\usepackage[pagebackref,breaklinks,colorlinks,allcolors=iccvblue]{hyperref}

\def\paperID{9238} 
\def\confName{ICCV}
\def\confYear{2025}

\title{PaintScene4D: Consistent 4D Scene Generation from Text Prompts}

\author{
\vspace{2mm} 
        Vinayak Gupta$^1$\hspace{16mm}
        Yunze Man$^2$ \hspace{16mm}
        Yu-Xiong Wang$^2$
    \vspace{2mm}
    \\
    \hspace{-3mm}
    \textsuperscript{1} Indian Institute of Technology, Madras
    \hspace{4mm}  
    \textsuperscript{2} University of Illinois Urbana-Champaign 
    \vspace{2mm}
    \\
    \hspace{-8mm}
    {\tt \href{https://paintscene4d.github.io/}{https://paintscene4d.github.io/}}
    }
    
\begin{document}
\twocolumn[{%
\renewcommand\twocolumn[1][]{#1}%
\maketitle\vspace{-1em}
\includegraphics[width=1\textwidth]{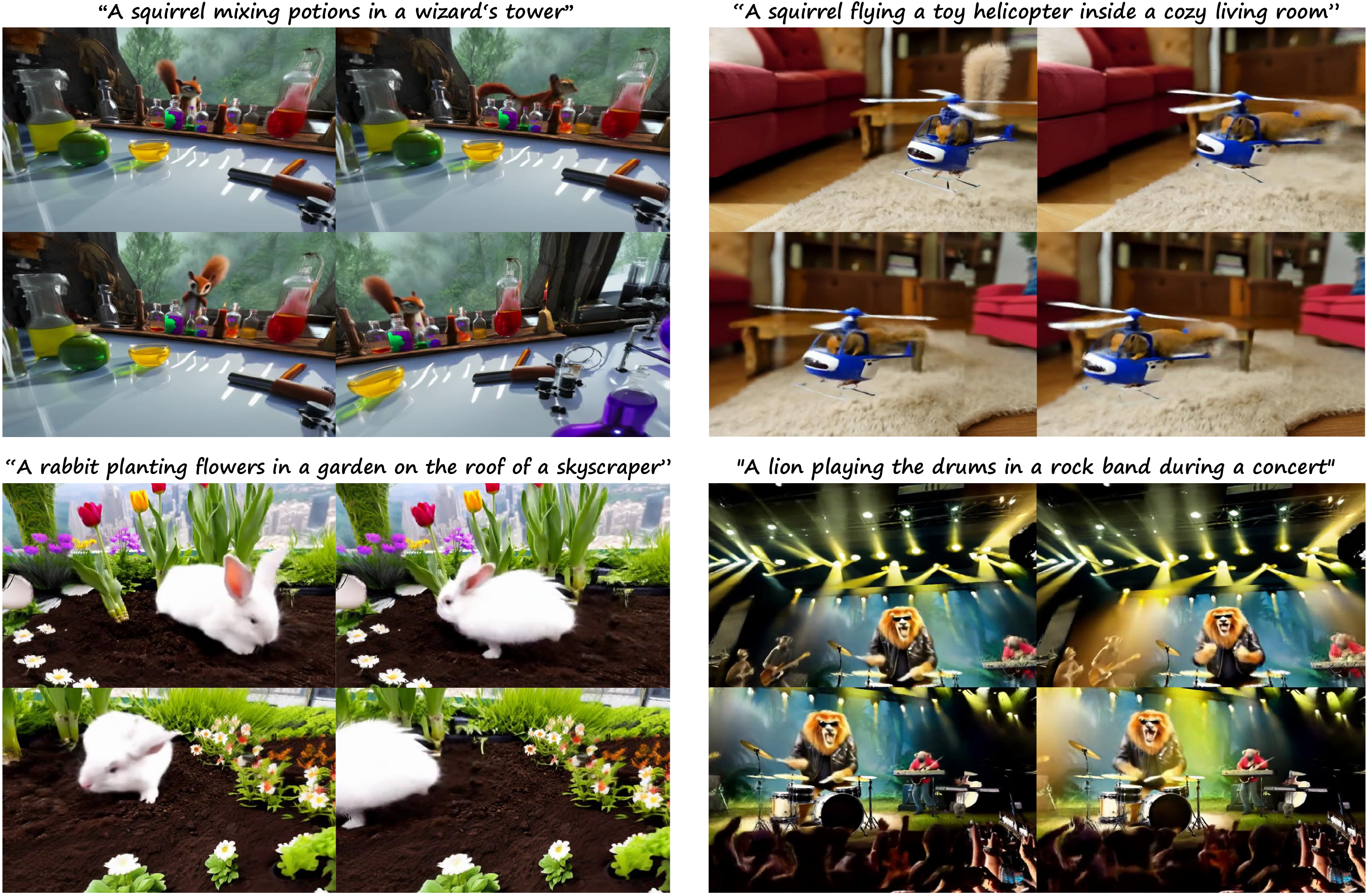}
\vspace{-6mm}
\captionof{figure}{\textbf{4D Text-to-Scene Generation.} Unlike prior methods that restrict text-to-4D generation to object-level reconstruction or text-to-video models lacking explicit camera control, our approach reconstructs full realistic 4D scenes that can be viewed from different trajectories, achieving via an efficient training-free architecture.}
\vspace{3mm}
\label{fig:teaser}
}]

\begin{abstract}
Recent advances in generative models have revolutionized 2D and 3D content creation, yet generating photorealistic 4D scenes remains a significant challenge. Existing methods typically rely on distilling knowledge from pre-trained 3D generative models, often fine-tuned on synthetic object datasets, resulting in scenes that are object-centric and lack photorealism. While text-to-video models can generate more realistic scenes with motion, they often struggle with spatial understanding and limited camera trajectory control. To address these limitations, we present PaintScene4D, a novel text-to-4D scene generation framework that departs from conventional 4D generative models in favor of a streamlined architecture that harnesses video generative models trained on diverse real-world datasets. Our model starts with the creation of a reference video and then employs a strategic camera trajectory control and a camera array selection module for novel view rendering. We introduce a progressive warping and inpainting strategy to ensure spatial and temporal consistency. Finally, we optimize novel-view videos using a dynamic renderer, enabling flexible camera control based on user preferences. Demonstrating the first training-free approach for 4D scene generation, PaintScene4D efficiently produces realistic and dynamic scenes viewed from arbitrary trajectories. The code will be made publicly available.
\end{abstract}    
\section{Introduction}

Generating dynamic 3D scenes from text descriptions, known as text-to-4D scene generation, represents one of the most challenging frontiers in vision and graphics. While recent advances have revolutionized our ability to create videos~\cite{modelscope,rombach2022high, singer2022make,saharia2022photorealistic} and static 3D content~\cite{liu2023zero, shi2023MVDream, liu2023syncdreamer,poole2022dreamfusion, lin2023magic3d, metzer2023latent, qian2023magic123}, the synthesis of temporally coherent and animated scenes remains a fundamental challenge. 

The complexity of 4D scene generation stems from several interconnected challenges. First, unlike static 3D generation, 4D scenes must maintain spatial and temporal coherence simultaneously, meaning that any generated motion must be physically plausible and semantically meaningful while preserving geometric structure over time. Second, the lack of large-scale 4D scene datasets has limited the development of robust generation methods, resulting in most existing approaches relying on object-centric data without rich dynamics of scenes. Third, the computational complexity of spatial-temporal training makes it difficult to achieve high-quality results within reasonable time constraints.

Current approaches to these challenges broadly fall into two categories, each with significant drawbacks. The first category extends static 3D generation methods~\cite{shi2023MVDream,liu2023zero,shi2023zero123pp} trained on object-centric datasets~\cite{deitke2023objaverse} to incorporate temporal dynamics~\cite{ling2023align,singer2023text,zheng2024unified,yin20234dgen,ren2023dreamgaussian4d,zhao2023animate124,jiang2023consistent4d,bahmani20244d}. These methods, while effective at maintaining geometric consistency, struggle with complex motion and often produce only subtle deformations and translations. The second category is text-to-video (T2V) models~\cite{guo2023animatediff,yang2024cogvideox} that lack explicit 3D understanding, resulting in spatial inconsistencies and geometric artifacts. Both methods require significant training time, and neither adequately addresses the fundamental challenge of generating coherent spatial-temporal 4D scenes.

To address these limitations, we present \textbf{\ourwork}, a novel \textit{training-free} framework that harnesses the strengths of T2V generation and 4D-aware neural rendering (note that the post-processing 4D Gaussian renderer is learnable. But it requires less than an hour and, in principle, can be achieved in a training-free manner.)
\textit{Our key insight} is that by using video generation as an initial prior and reconstructing the 3D scene through a \textit{progressive warping and inpainting technique}, we can maintain spatial-temporal consistency while enabling complex motion generation. Specifically, our method first generates a base video using a pre-trained T2V model to provide rich motion priors. We then construct a ``web of cameras'' around the scene by warping the frames to novel viewpoints with a minimum-overlapping viewpoint selection. We propose the progressive warping module (PWM) and the consistent inpainting module (CIM), allowing us to determine an optimal sequence for warping and inpainting and build a consistent multi-view representation of the dynamic scene, \textit{without} requiring explicit 3D supervision or costly optimization.

The effectiveness of \ourwork is demonstrated through extensive empirical contributions. As shown in Figure~\ref{fig:teaser}, our method achieves state-of-the-art results in text-to-4D scene generation, producing visually compelling results that maintain spatial-temporal consistency. The generated scenes exhibit complex motion while preserving geometric structure across multiple viewpoints. Notably, our framework reduces computational requirements significantly due to its training-free manner, generating high-quality 4D content in approximately 2.2 hours on a single A100 GPU, a substantial improvement over existing methods~\cite{bahmani20244d, zheng2024unified} that often require 10+ hours. Through extensive ablation studies, we demonstrate the superiority of our approach across various metrics, including temporal consistency, motion complexity, and rendering quality. \textit{Our method also offers notable flexibility, allowing users to edit existing videos or specify custom trajectories during inference.}

Our main contributions can be summarized as follows.
\begin{itemize}
    \item We propose a novel modularized and training-free framework for text-to-4D scene generation that effectively distills video generation prior to 4D-aware neural rendering.
    \item We introduce a progressive warping and inpainting technique, and demonstrate key technical designs to achieve high-quality results with the combination of video generation and inpainting methods. 
    \item We perform a comprehensive evaluation and analysis of \ourwork, showing leading results in the generation of 4D scenes, with significantly reduced computational requirements and enhanced camera control options.
\end{itemize}

\label{sec:introduction}

\begin{figure*}[!t]
  \vspace{-4mm}
  \includegraphics[width=\textwidth]{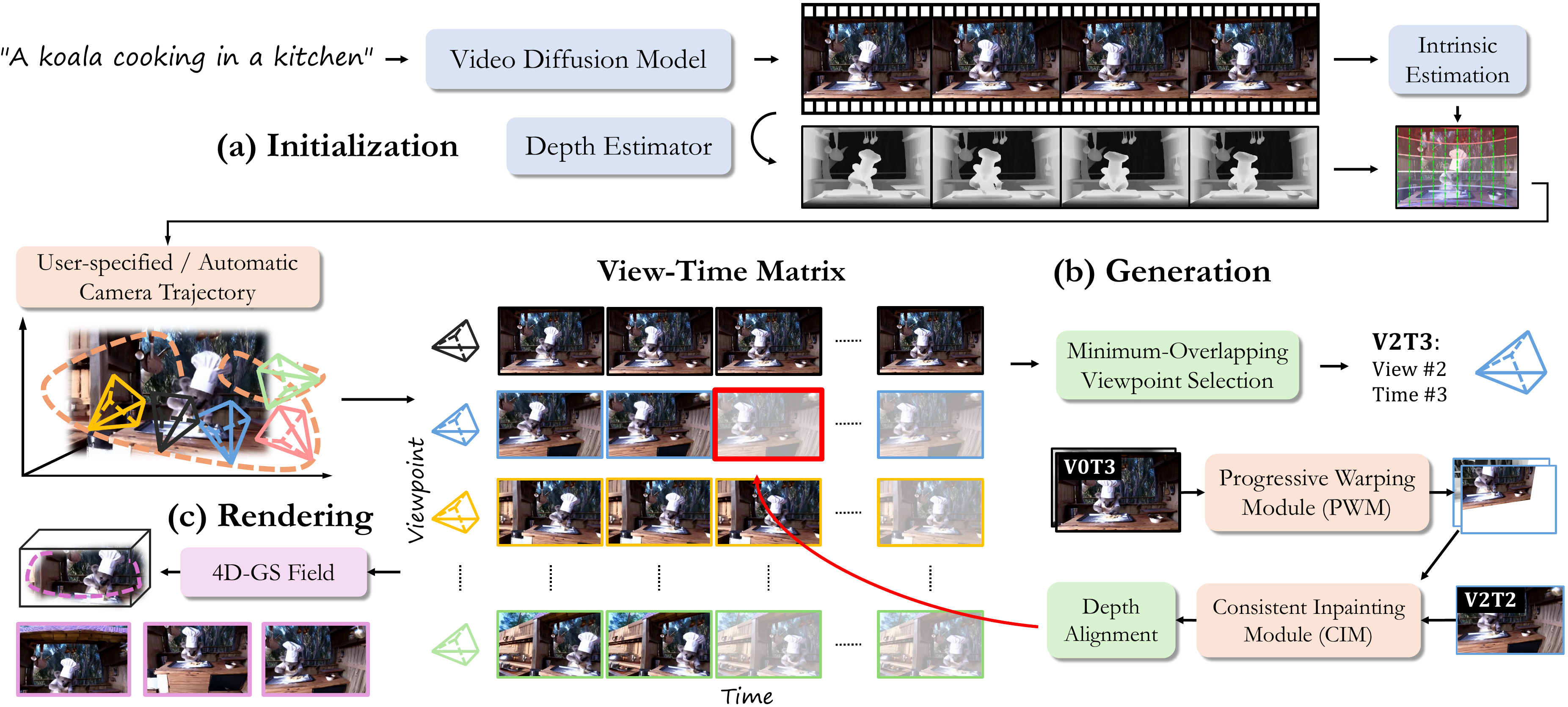}
  \vspace{-6mm}
  \caption{\textbf{Method Overview.} Our approach consists of three stages. First, we initialize the 4D scene using a diffusion prior to establish scene content and motion, estimate depth maps for each video frame, and initialize camera trajectory (extrinsics) and intrinsics for subsequent warping. In the second stage, we perform sequential warping and inpainting from the first timestamp. To ensure spatial and temporal coherence, our consistent inpainting module mitigates artifacts and aligns depth maps, preventing error accumulation. Finally, the generated view-time matrix is used to render novel views along user-defined camera trajectories, allowing for explicit camera control.}
  \label{fig:modelarch}
  \vspace{-3mm}
\end{figure*}

\section{Related Work}
\label{sec:related_work}

\mypar{Text-to-3D Generation.} Text-to-3D generation has evolved significantly over the past decades. Initial approaches rely on rule-based systems that parse text inputs into semantic representations for scene generation using object databases~\cite{adorni1983natural,coyne2001wordseye,chang2014learning}. The field has advanced substantially with the introduction of data-driven approaches that leverage multimodal datasets~\cite{chen2018text2shape} and pre-trained models like CLIP~\cite{radford2021learning}, enabling more sophisticated manipulation of 3D meshes~\cite{jetchev2021clipmatrix, gao2023textdeformer} or radiance fields~\cite{wang2022clip}. This progress has led to the development of methods utilizing CLIP-based supervision for comprehensive 3D scene synthesis~\cite{jain2022zero, sanghi2022clip}, which subsequently evolves into techniques that optimize meshes and radiance fields through Score Distillation Sampling (SDS)~\cite{poole2022dreamfusion, wang2023prolificdreamer, lin2022magic3d}. The introduction of multi-view-aware diffusion models has further enhanced the quality of generated 3D structures~\cite{lin2023consistent123, liu2023zero, shi2023MVDream}. Parallel developments in diffusion and transformer architectures have enabled advanced image-to-3D conversion for novel view synthesis~\cite{chan2023generative, tang2023make, gu2023nerfdiff, liu2023syncdreamer, yoo2023dreamsparse, tewari2023diffusion, qian2023magic123}. These approaches primarily address object-level reconstruction.

Recent advances in text-to-3D scene generation have introduced innovative approaches to address scene-level complexity. Text2Room~\cite{hollein2023text2room} proposes a warping and inpainting methodology for scene creation, while Text2NeRF~\cite{zhang2024text2nerf} shifts away from mesh-based reconstruction to utilize radiance fields as scene generation priors. Subsequent work~\cite{zhang20243d} expands the capabilities to support general 3D scene generation with arbitrary 6 degree-of-freedom camera trajectories. However, these approaches remain limited to static scenes, lacking the ability to incorporate motion, a crucial element for dynamic environments.

\vspace{1mm}
\mypar{Object-centric Text-to-4D Generation.} The extension from 3D to 4D scene generation introduces significant additional complexity. MAV3D~\cite{singer2023text} pioneers this direction by introducing a dynamic neural radiation field (NeRF) representation using HexPlane~\cite{cao2023hexplane} and a video-based SDS loss. Dream-in-4D~\cite{zheng2024unified} employs a dynamic NeRF, organizing text-to-4D generation into distinct static and dynamic phases. Similarly, 4D-fy~\cite{bahmani20244d} introduces a hybrid representation that combines static and dynamic voxels with SDS loss functions~\cite{poole2022dreamfusion,shi2023MVDream,luo2023videofusion}. Additionally, AYG~\cite{ling2024align} achieves dynamic rendering through the application of a dynamic network to 3D Gaussian splatting. Recent developments have focused on decomposing and controlling motion generation. TC4D~\cite{bahmani2025tc4d} separates motion into global trajectories with user-defined global paths and local motion generated in segments. Comp4D~\cite{xu2024comp4d} employs the large language model (LLM) to decompose the prompts into entities, generating 4D objects with LLM-derived trajectories. These approaches focus on object-level reconstruction, limiting their broader applicability. 

\vspace{1mm}
\mypar{Scene-centric Text-to-4D Generation.} A notable departure from this trend is 4Real~\cite{yu20244real}, which circumvents multi-view generative models by leveraging video generative models trained on large-scale datasets. VividDream~\cite{lee2024vividdream} generates explorable 4D scenes with ambient dynamics with pre-trained video diffusion and inpainting modules. Another concurrent work CAT4D~\cite{wu2024cat4d} generates 4D dynamic scenes from monocular videos by a pre-trained multi-view video diffusion model and a deformable 3D Gaussian representation for novel view synthesis. Our approach differs from previous methods by introducing a training-free framework, generating 4D scenes that accurately capture both the geometric structure of real-world environments while providing greater control over camera movement and rendering. Additionally, our method enables the rendering of a 4D scene in just 2 to 3 hours, making it computationally efficient and practical for real-world applications.

\section{Method}
\label{sec:method}

\textbf{Overview.} In this work, we present \ourwork, a novel framework designed to generate 4D dynamic scenes from textual inputs. Our approach begins with an initial diffusion-generated video that serves as both a scene and a motion reference. Using this video as input, we employ a depth estimation model to derive depth maps from each frame, allowing us to progressively construct a spatial representation of the scene. To create a comprehensive multi-camera view of the scene, we develop a progressive warping module (PWM) where regions missing due to occlusion or perspective changes during the warping process are progressively filled in using a spatially consistent inpainting module (CIM). For each subsequent frame, our approach reuses inpainted data from prior timestamps for continuity and only fills in new, unobserved areas. Once we have constructed a network of cameras, where each camera captures all frames over time, we employ a 4D rendering algorithm to reconstruct the scene and generate novel viewpoints. This entire methodology is described in Figure \ref{fig:modelarch}.

\begin{figure*}[!t]
  \vspace{-1mm}
  \includegraphics[width=\textwidth]{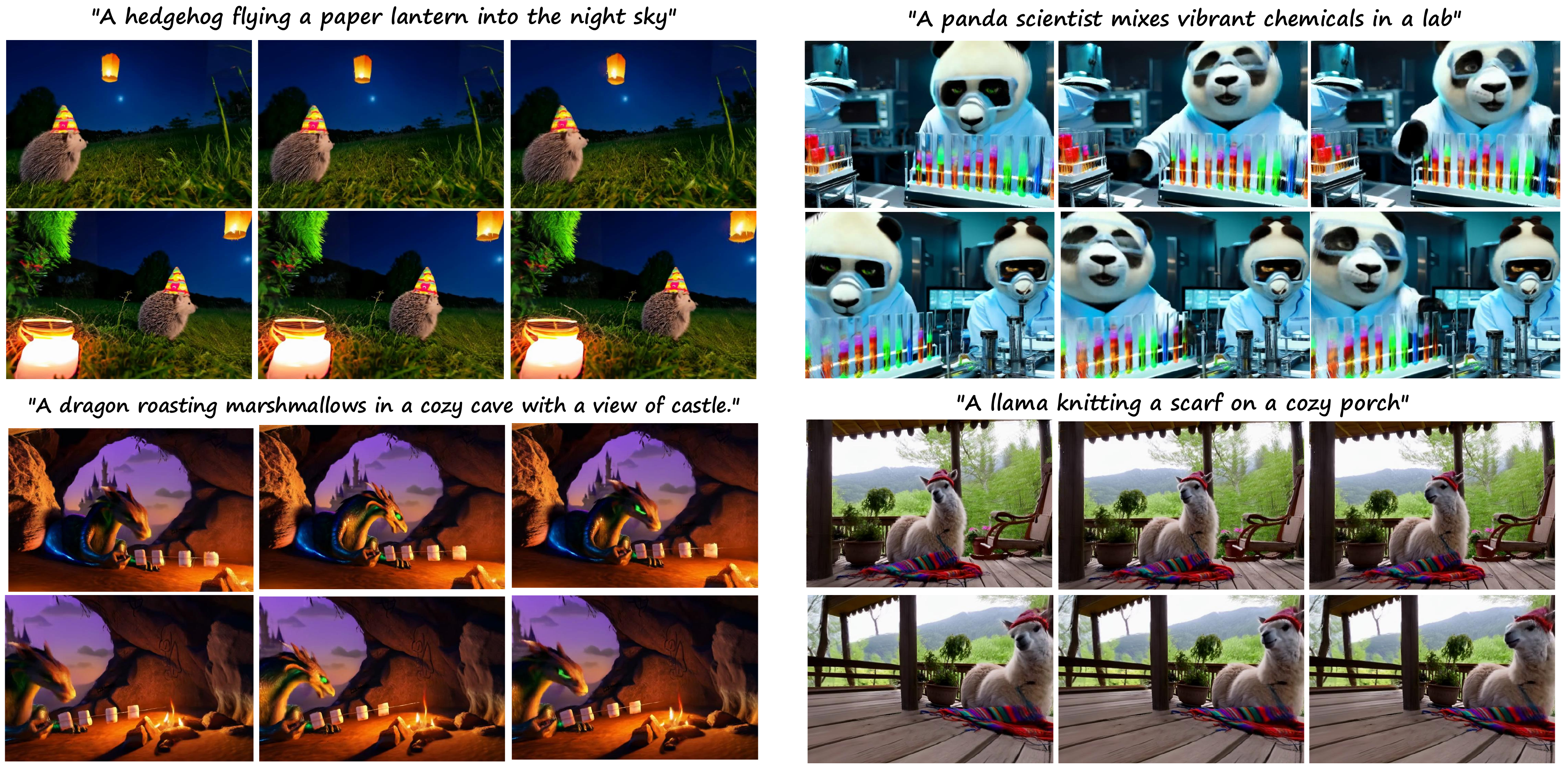}
  \vspace{-7.5mm}
  \caption{\textbf{Gallery of Results.} \ourwork successfully generates 4D scenes that maintain view- and temporal-coherence. The horizontal axis represents the time; the vertical axis represents different viewpoints. More visualizations are provided in the supplementary materials.}
  \vspace{-5mm}
  \label{fig:gallery-main}
\end{figure*}

\subsection{Scene Initialization}
\label{sec:scene_init}
\mypar{Reference Video Generation and Depth Estimation.} To generate the initial scene content from an input prompt \( t \), we start by applying a pre-trained video diffusion model, \( f_{d} \), conditioned on \( t \) to create an initial video \( V_0 = f_{d}(\epsilon \mid t) \), where \( \epsilon \) is random Gaussian noise. Given our approach requires the video to be captured using a stationary, non-moving camera, we enhance the user-defined prompt with additional descriptors, such as ``The camera remains stationary, with a fixed frame, stable composition, and no shifts.'' This added specificity ensures that the output video aligns with our fixed-camera requirement. As \( V_0 \) does not contain inherent geometric depth information, we integrate a video depth estimation model \( f_{e} \) to obtain this depth data, generating depth maps \( D_0 = f_{e}(V_0) \). The video frames in \( V_0 \) paired with the corresponding depth maps \( D_0 \) serve as the basis for initializing the 4D scene. 

\vspace{1mm}
\mypar{Camera Trajectory.} To support the intended trajectory of the final rendered output, we establish a network of virtual cameras to match the user’s desired camera path. These cameras represent a structured arrangement of views that form the backbone for the construction of the 4D scene. Given that our framework incorporates warping operations, it is imperative to obtain accurate intrinsics of the camera parameters in the generated videos. To address this, we employ a pre-trained model, Perspective Field~\cite{jin2023perspective}, to calculate the intrinsic matrix based on the video frames provided. 

\subsection{Progressive Warping Module (PWM)}

Given the absence of multi-view supervision, directly employing a single-view video \( V_0 \) and its depth maps \( D_0 \) to train a 4D radiance field can lead to issues of overfitting and geometric ambiguity. To address this, we apply a depth image-based rendering technique (DIBR~\cite{fehn2004depth}) to establish a network of virtual cameras around the initial view. Specifically, for each pixel \( p \) in \( I^t_i \) and its corresponding depth \( z \) in \( D^t_i \), we compute its transformed coordinates \( p_{i \rightarrow j} \) and depth \( z_{i \rightarrow j} \) for a neighboring viewpoint \( j \) as follows:

{\small
  \setlength{\abovedisplayskip}{-6pt}
  \setlength{\belowdisplayskip}{6pt}
  \setlength{\abovedisplayshortskip}{0pt}
  \setlength{\belowdisplayshortskip}{3pt}
 \begin{align}
    \left[p_{i \rightarrow j}, z_{i \rightarrow j}\right]^{T} = \mathbf{K}\mathbf{P}_{j}\mathbf{P}_{i}^{-1}\mathbf{K}^{-1}\left[p, z\right]^{T},\label{eq:dibr}
\end{align}
}%
where \( \mathbf{K} \), \( \mathbf{P}_{i} \) and \( \mathbf{P}_{j} \) are the intrinsic matrix, camera pose for view \( i \) and view \( j \), respectively, and \( I^t_i \) represents the image at timestamp $t$ of viewpoint $i$. Following the transformation, we fill the missing or occluded regions in the newly warped views with inpainting. Our experiments reveal that the inpainting diffusion-based prior yields higher-quality results when the inpainted regions are larger. Therefore, for each view, we select the \textit{farthest available viewpoint with minimal overlap}, warp the current frame to this viewpoint, and apply inpainting as necessary. Large occlusions are filled using a 2D diffusion-based prior, while smaller gaps are addressed with Telea-based inpainting~\cite{telea2004image}. Refer to Sec.~\ref{sec:ablation} for quantitative and qualitative ablation studies on this.

Our warping process begins at the first timestamp, progressively warping and inpainting frames across all views before proceeding to subsequent timestamps. For the first timestamp, we start with a base view \( I^0_0 \), warp it to a neighboring viewpoint \( I^0_1 \), and paint any missing regions. To ensure spatial consistency, we integrate both the original (\( I^0_0 \)) and newly warped frames (\( I^0_1 \)) for further warping (\eg, \( I^0_2 \), \( I^0_3 \)). This approach ensures that any inpainted content in \( I^0_1 \) is preserved in subsequent viewpoints (\( I^0_2 \), \( I^0_3 \), etc.), maintaining coherence throughout the scene.

\vspace{1mm}
\mypar{Depth Alignment.} To transform a 2D image \( I \) into a 3D representation, we first estimate the depth for each pixel. Accurate integration of both new and existing content requires precise depth alignment, ensuring that similar elements in the scene, such as walls or furniture, appear at consistent depths across views. Directly projecting the predicted depth often results in abrupt transitions and geometric discontinuities due to inconsistent scale across viewpoints. To address this, we apply a depth alignment procedure inspired by~\citet{liu2021infinite}, which refines the depth through scale and shift optimization. Specifically, we optimize scale \(\gamma\) and shift \(\beta\) parameters \(\gamma, \beta \in \mathbb{R}\) by minimizing the difference between the predicted depth $\hat{d}$ and the rendered depth ${d}$ in the least-squares sense.

{\small
  \setlength{\abovedisplayskip}{-6pt}
  \setlength{\belowdisplayskip}{6pt}
  \setlength{\abovedisplayshortskip}{0pt}
  \setlength{\belowdisplayshortskip}{3pt}
 \begin{align}
    \min_{\gamma, \beta} \quad \left \lVert m \odot \left( {\gamma}{\hat{d}} + \beta - {d} \right) \right \rVert^2,\label{eq:align-step-2}
\end{align}
}%
Where mask \( m \) excludes unobserved pixels from the alignment. Additionally, depth estimation models may fail to accurately resolve depth at object boundaries, often yielding smooth transitions where abrupt changes are expected. This issue affects the overall warping quality, resulting in artifacts such as trailing patterns within occluded regions. To address this, we apply bilateral filtering to sharpen the depth boundaries, enhancing inpainting performance. Additional details are provided in the supplementary material.

\subsection{Consistent Inpainting Module (CIM)}

Upon completing the warping and inpainting for the first timestamp, we proceed to apply these operations sequentially across subsequent timestamps. However, directly extending the same approach to each timestamp independently can lead to temporal inconsistencies. This is due to the inherent variability in 2D diffusion-based inpainting, which may produce differing results for the same regions across different timestamps. To address this, we impose temporal consistency by ensuring that background regions remain visually coherent across frames. Specifically, we require that overlapping regions across timestamps exhibit similar content, especially in the background areas.

\vspace{1mm}
\mypar{Foreground and Background Separation.} After the inpainting process, we use a segmentation model to separate the foreground and background regions within each frame. For regions that contain significant occlusions, especially large missing areas in the background, we incorporate content from previous timestamps to fill these areas. This approach maintains temporal continuity by sourcing background information from earlier frames. For holes near the foreground boundary, we determine the inpainting source based on the background or foreground status of the corresponding region in prior timestamps.  If a boundary region classified as background in the current frame aligns with a background area in previous timestamps, we inpaint it using information from the earlier frame. Conversely, if the region is identified as part of the foreground in prior frames, we apply the 2D diffusion model for inpainting. This selective inpainting strategy allows us to maintain coherence across timestamps while appropriately filling areas based on temporal foreground and background information.

\begin{figure*}[!t]
  \includegraphics[width=\textwidth]{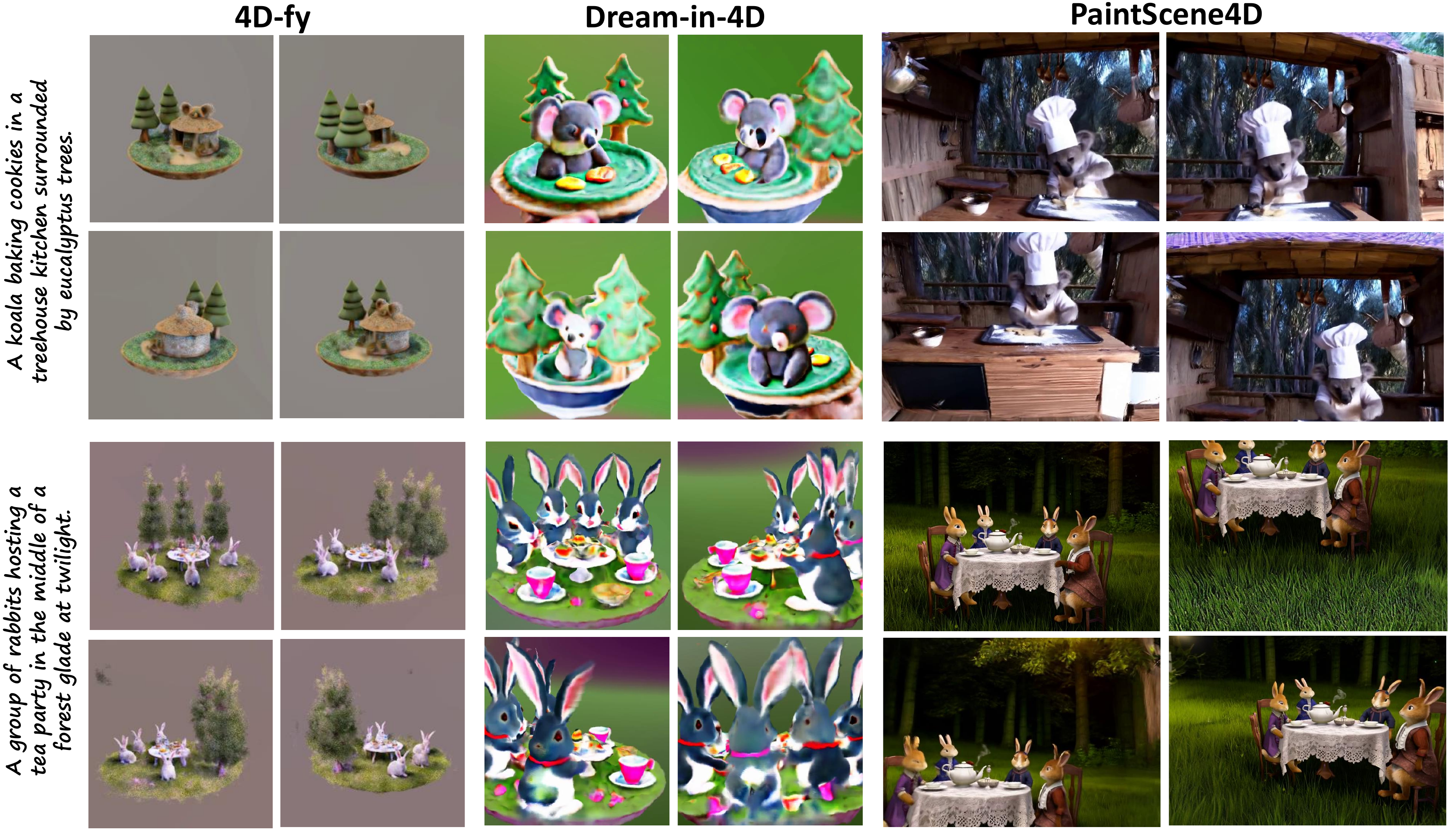}
  \vspace{-6.5mm}
  \caption{\textbf{Comparisons with state-of-the-art text-to-4D generation methods.} While both baseline methods produce scenes that broadly align with the text prompts, they lack essential fine details. Specifically, 4D-fy shows minimal motion and limited detail, whereas Dream-in-4D captures dynamics more effectively but produces stylized, cartoon-like renderings. In contrast, our method synthesizes photorealistic 4D scenes that faithfully follow the input text prompt while presenting significant, realistic dynamics within the scene.}
  \vspace{-6mm}
  \label{fig:exp1-main}
\end{figure*}


\begin{figure}[!t]
  \includegraphics[width=\linewidth]{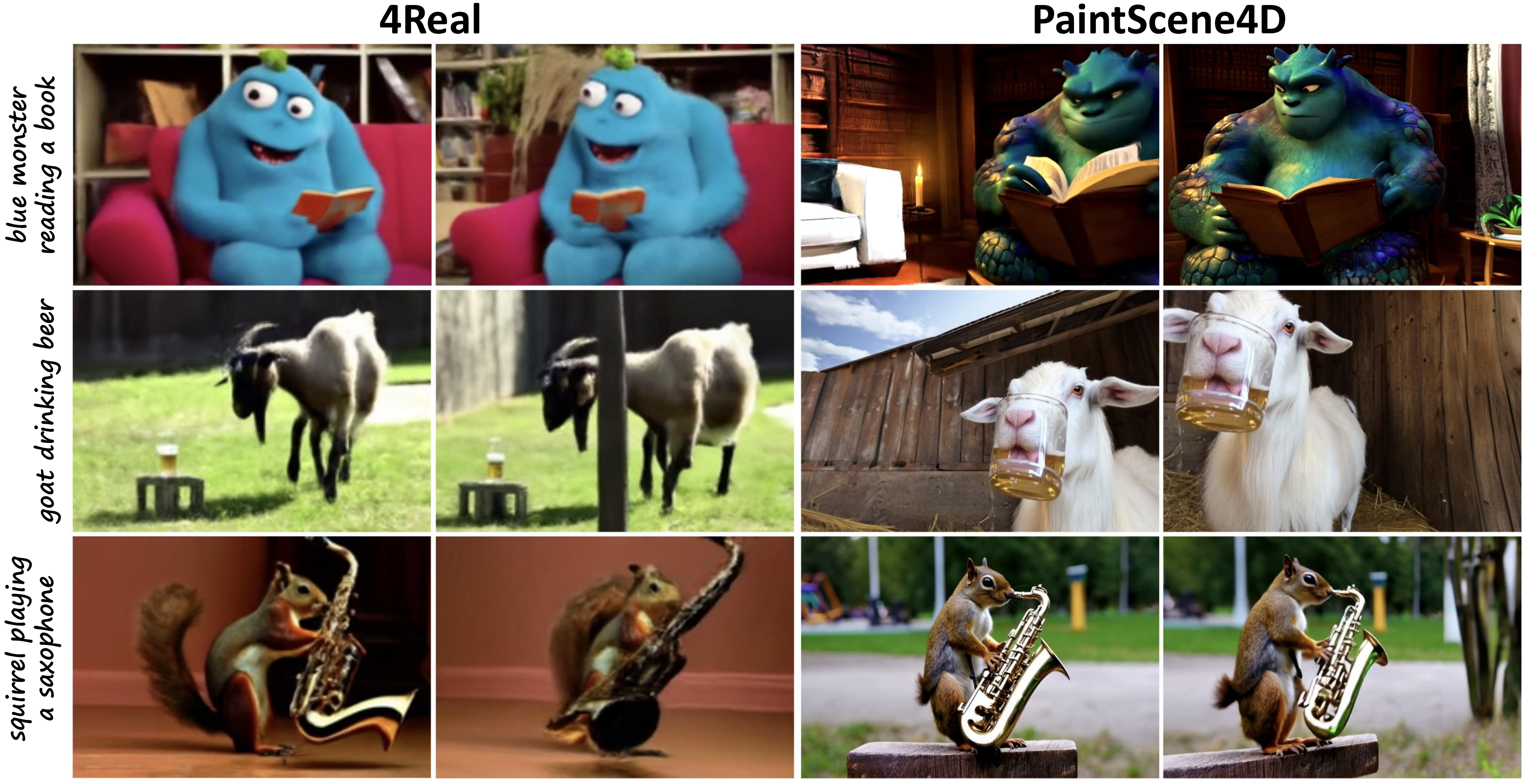}
  \vspace{-6.5mm}
  \caption{\textbf{Comparison against 4Real.}~\cite{yu20244real} We demonstrate that our method produces more dynamics, larger scene coverage and better video-text alignment, and overall realism scenes.}
  \vspace{-5mm}
  \label{fig:4real_comparison}
\end{figure}

\begin{figure}[!t]
  \includegraphics[width=\linewidth]{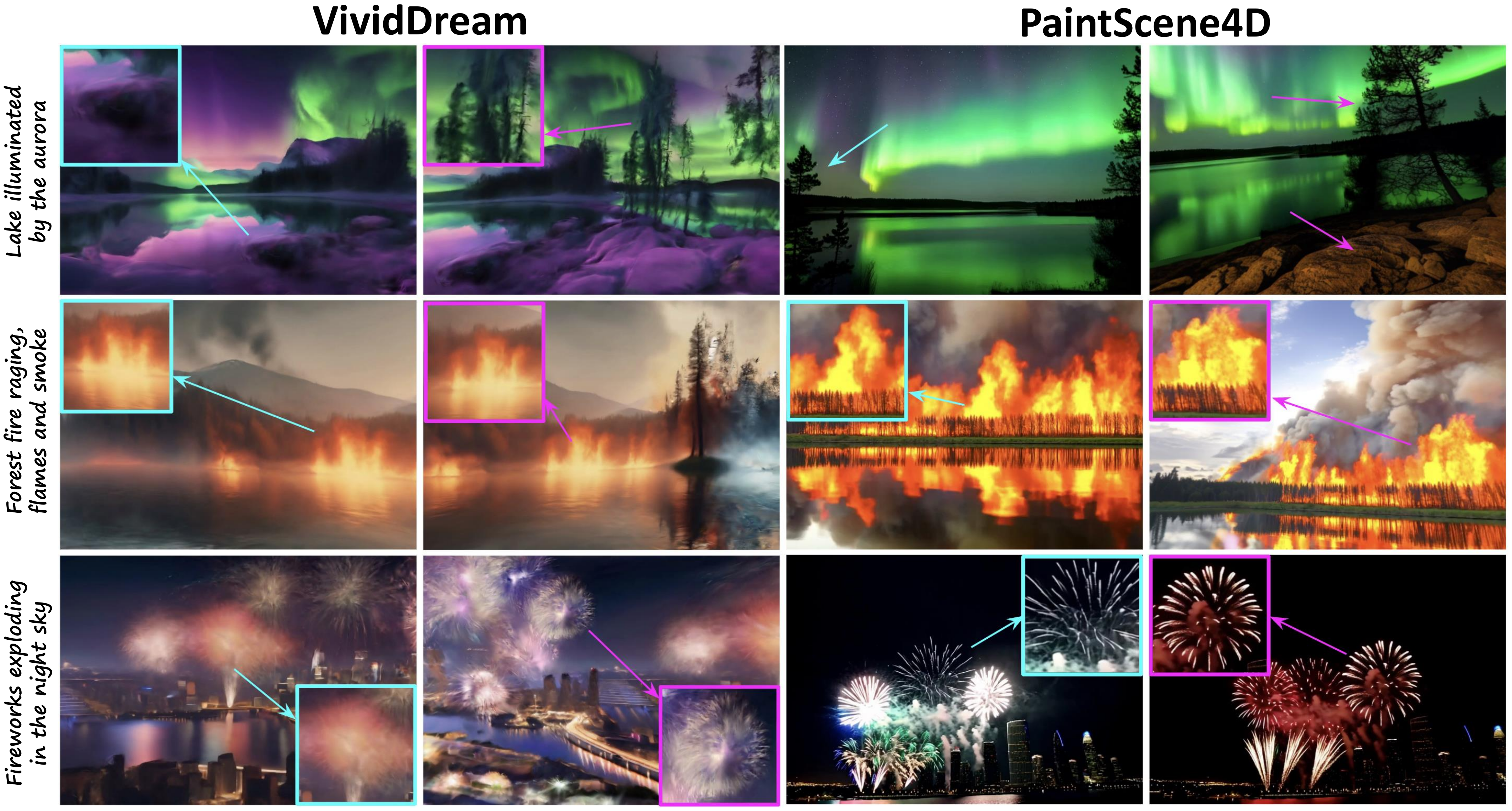}
  \vspace{-6.5mm}
  \caption{\textbf{Comparison against VividDream.}~\cite{lee2024vividdream} PaintScene4D demonstrates superior performance compared to VividDream in terms of higher quality (row 1), more dynamics (row 2), and higher motion and overall realism (row 3).}
  \vspace{-5mm}
  \label{fig:vividdream_comparison}
\end{figure}


\subsection{Training and Optimization}
After performing all warping and inpainting operations across views and timestamps, we establish a comprehensive camera network, where each camera contains video frames captured from its respective viewpoint. Importantly, this multi-view setup is constructed without the need for model-specific training. Using this multi-view spatial information and temporal dynamics, we employ a 4D Gaussian rendering approach~\cite{wu20244d} to synthesize novel perspectives of the scene. The renderer takes Gaussian parameters, along with the timestamp, and computes the timestamp-conditioned deformation of these parameters. This approach enables continuous modeling of deformation, facilitating smooth interpolation between timestamps during novel view synthesis. At test time, any desired viewpoint and timestamp can be selected to generate a novel view.

\section{Experiments}
\label{sec:analysis}

In this section, we introduce the baselines to compare with, the evaluation metrics, qualitative and quantitative results, and the analysis. The implementation details and additional analysis are presented in the supplementary material.

\begin{figure*}[!t]
  \vspace{-2mm}
  \includegraphics[width=\textwidth]{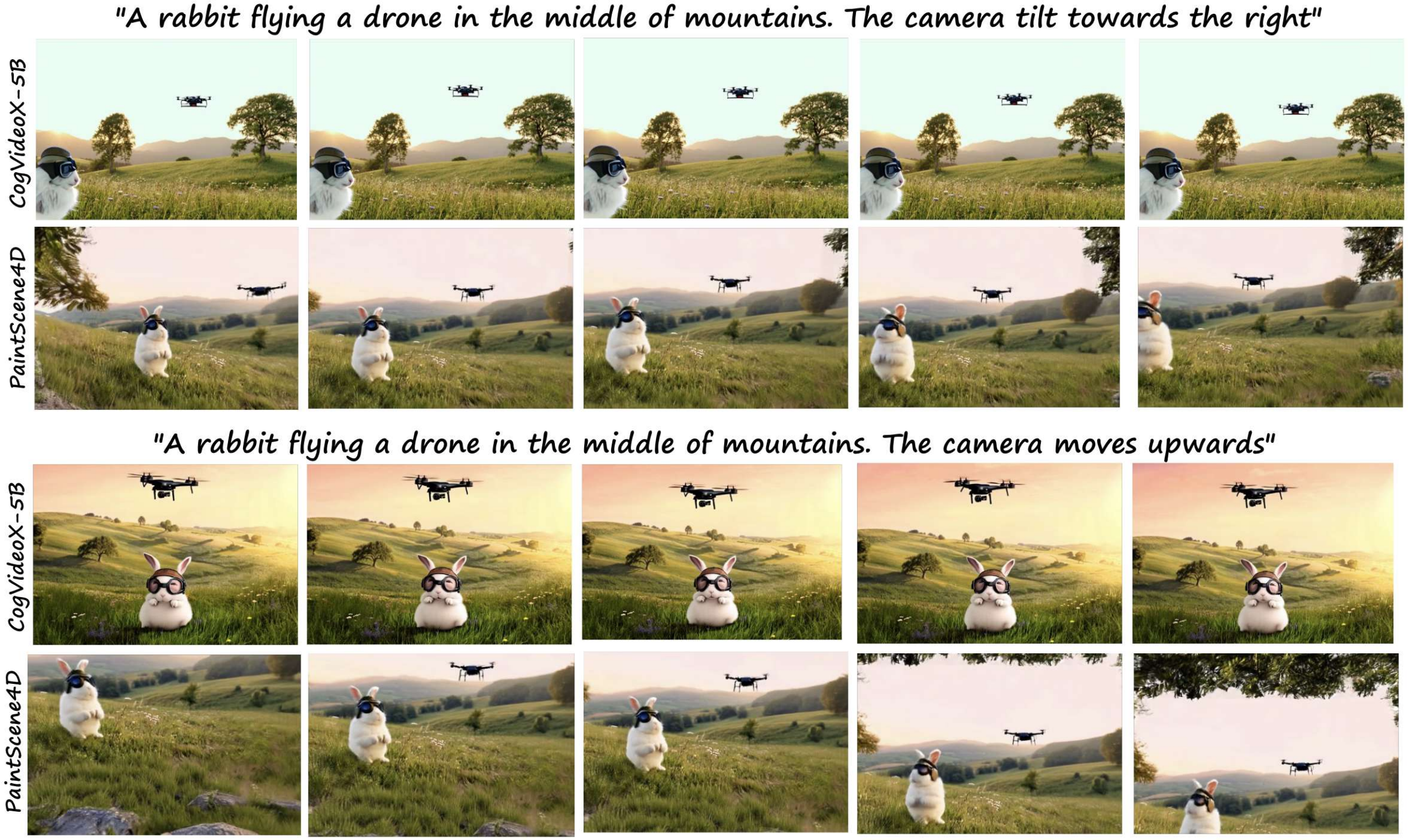}
  \vspace{-7mm}
  \caption{\textbf{Camera Control.} \ourwork shows strong explicit camera control capabilities. To guide T2V models (\textit{e.g.,} CogVideoX~\cite{yang2024cogvideox}), we append camera motion directives to the text prompt such as ``The camera tilts to the right / upwards.'' However, due to the implicit handling of camera motion, T2V often fails to generate precise or controllable camera movements. Our approach, once trained, allows for flexible camera trajectories within the bounds of the input cameras, achieving precise and repeatable control over camera movements.}
  \vspace{-5mm}
  \label{fig:exp2-main}
\end{figure*}

\begin{figure}[!t]
  \includegraphics[width=0.48\textwidth]{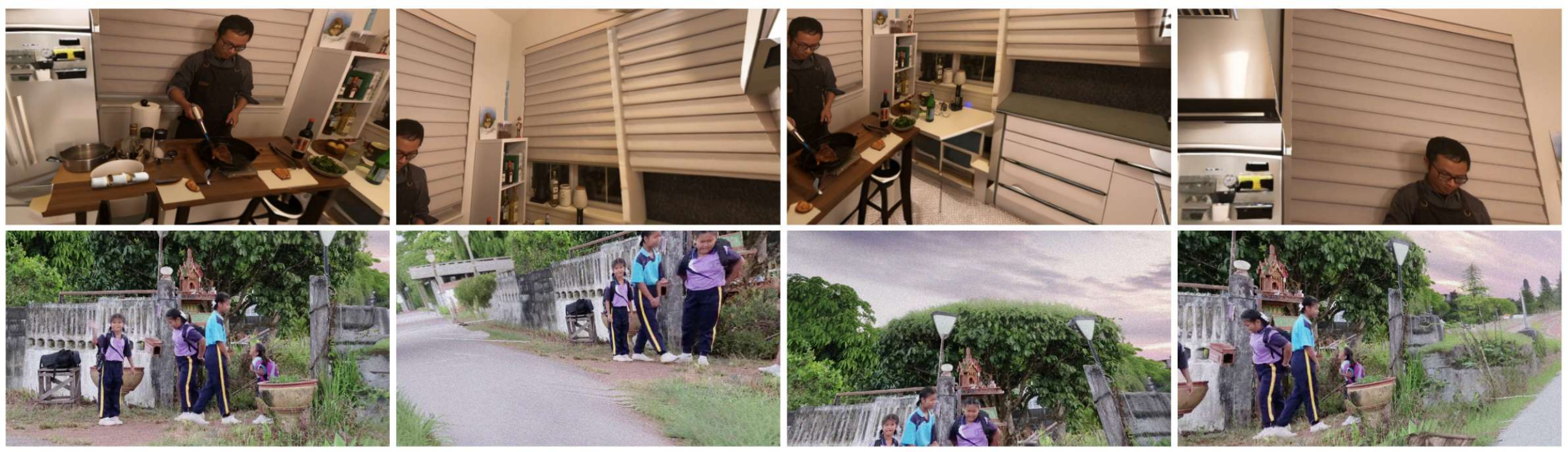}
  \vspace{-7mm}
  \caption{\textbf{Qualitative Results on real-world videos}. Our work can produce realistic and coherent results with real-world video inputs, rather than limited to generated videos from T2V models.}
  \label{fig:real-world}
  \vspace{-5mm}
\end{figure}

\subsection{Baselines and Evaluation Metrics}

In the absence of open-source implementations for text-to-4D scene-level generation, we benchmark our approach against state-of-the-art text-to-4D object-level generation methods, namely 4Dfy~\cite{bahmani20244d} and Dream-in-4D~\cite{zheng2024unified}, across a varied set of 20 prompts. We also compare against closed-source scene-level models~\cite{yu20244real,lee2024vividdream} with examples shown in their paper using the same text prompts. To assess the effectiveness of our proposed approach, we use the CLIP Score~\cite{radford2021learning} alongside a structured user study. More visualization and comparisons with other closed-source models~\cite{singer2023text,bahmani2025tc4d} are included in the supplementary material. 

\vspace{1mm}
\mypar{CLIP Score.} The CLIP score~\cite{park2021benchmark} assesses the alignment between textual prompts and visual contents by calculating the cosine similarity between their embeddings~\cite{radford2021learning}. Scored between 0 and 100, a higher value indicates a closer match. We compute CLIP scores by evaluating each frame with CLIP ViT-B/32 and averaging the scores across all frames and prompts for consistency.

\vspace{1mm}
\mypar{User Study.} A comprehensive user study is conducted using Google Forms, involving 30 evaluators per video pair. Each evaluator receives three anonymized videos, each capturing a dynamic scene from a camera moving along a circular trajectory. The videos are accompanied by the original text prompt. The evaluators were asked to rate their preferences based on four criteria: motion realism, video-text alignment, high dynamicity, and general realism.

\subsection{Text-to-4D Generation}

\begin{table}[!t]
    \caption{\textbf{Quantitative results.} We compare our method against object-level methods~\cite{bahmani20244d,zheng2024unified} (above the dotted lines) and against scene-level methods~\cite{yu20244real} (below the dotted lines). The methods are evaluated in terms of CLIP score and human preference, including motion realism (MR), video-text alignment (VTA), high dynamicity (HR), general realism (GR), and overall preference. The reported human preference is the percentage of users who voted for the respective method in a head-to-head comparison.}
    \vspace{-6mm}
    \label{tab:quanresults}
    \begin{center}
    \resizebox{\columnwidth}{!}{
    \begin{tabular}{lc|cccc|c}
        \toprule
         &  & \multicolumn{5}{c}{\textit{Human Preference}}\\
        \textit{Method} & CLIP$\uparrow$ & MR$\uparrow$ & VTA$\uparrow$ & HR$\uparrow$ & GR$\uparrow$ & Overall$\uparrow$ \\\midrule
        4D-fy~\cite{bahmani20244d} & 31.8 & 2\% & 11\% & 5\% & 7\% & 7\%\\
        Dream-in-4D~\cite{zheng2024unified} & 28.1 & 13\% & 14\% & 17\% & 2\% & 11\%\\
        \textbf{PaintScene4D} & \textbf{36.0} & \textbf{85\%} & \textbf{75\%} & \textbf{78\%} & \textbf{91\%} & \textbf{82\%}\\
        \hdashline
        4Real~\cite{yu20244real} & 33.7 & \textbf{59\%} & 42\% & 19\% & 39\% & 34\%\\
        \textbf{PaintScene4D} & \textbf{35.5} & 41\% & \textbf{58\%} & \textbf{81\%} & \textbf{61\%} & \textbf{66\%}\\
        \bottomrule
    \end{tabular}}
    \end{center}
    \vspace{-9mm}
\end{table}

\mypar{Qualitative Results.} In Figure~\ref{fig:exp1-main}, we visualize spatio-temporal renderings produced by our method compared to baselines. Although all approaches are capable of synthesizing 4D scenes, baselines focus on object-level renderings and lack fine spatial details. Our approach, by contrast, generates scene-level 4D reconstructions in a significantly reduced time, producing realistic renderings. Notably, 4D-fy struggles to model realistic motion, while Dream-in-4D produces cartoonish effects that diminish realism. 

We also compare our approach with closed-source scene generation models, 4Real and VividDream, as shown in Figure~\ref{fig:4real_comparison} and Figure~\ref{fig:vividdream_comparison}. Our observations indicate that 4Real tends to generate outputs with a more cartoon-like appearance and limited scene dynamics. Additionally, its resolution is restricted due to constraints imposed by SDS-based optimization. Similarly, while VividDream produces more realistic results, it struggles to capture the level of dynamic motion typically observed in real-world scenarios. In contrast, our method achieves high photorealistic quality across both spatial and temporal dimensions. A gallery showcasing our results is presented in Figure~\ref{fig:gallery-main}.

\vspace{1mm}
\mypar{Quantitative Results.} The CLIP score and the user study are reported in Table~\ref{tab:quanresults}. In both object- and scene-level comparisons, our method outperforms all baseline models. The evaluations show a statistically significant preference for \ourwork due to its higher motion realism, photorealistic rendering of both the foreground and background, overall realism, and better video-text alignment. 

\begin{figure}[!t]
  \includegraphics[width=\linewidth]{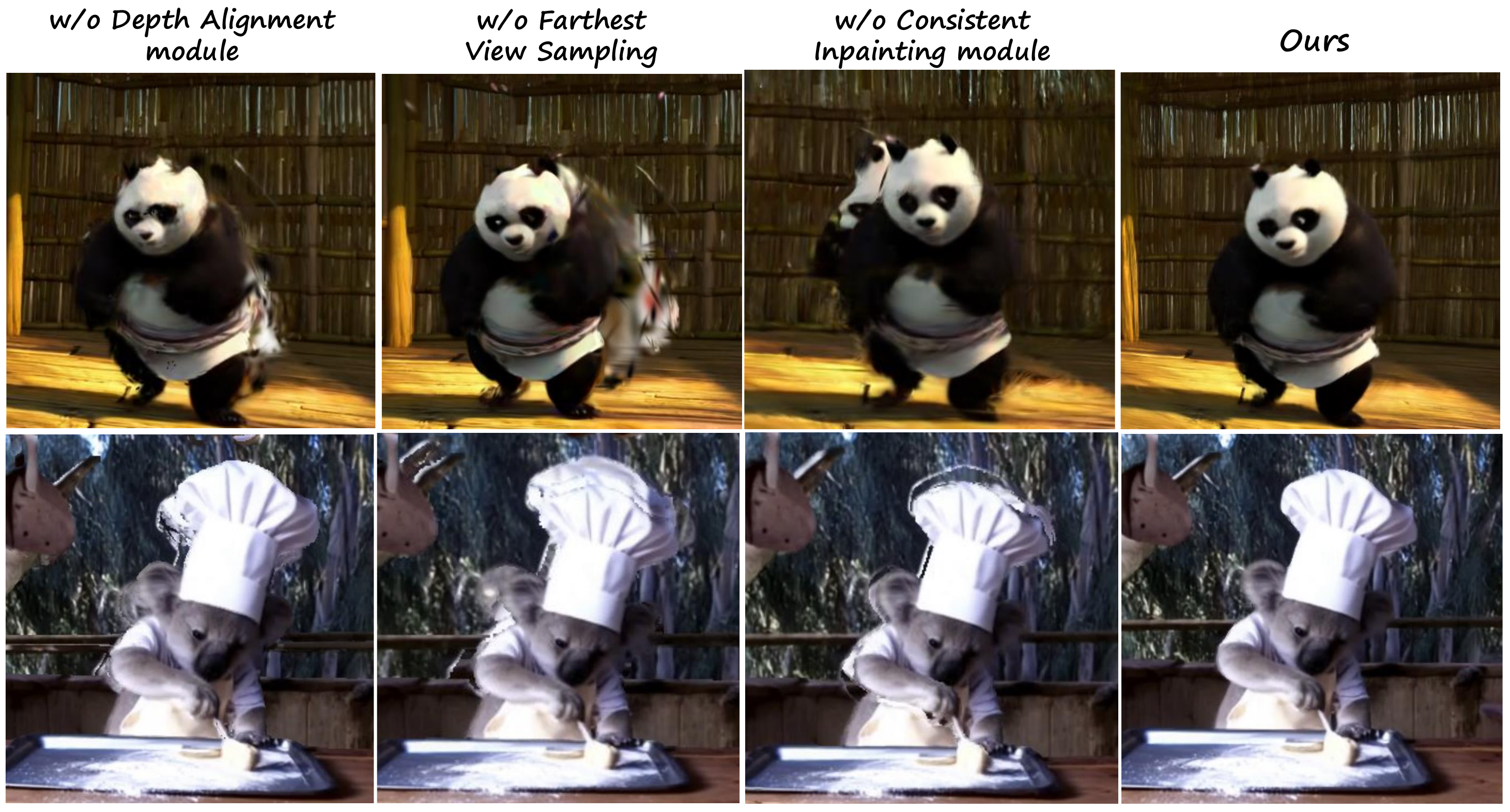}
  \vspace{-7mm}
  \caption{\textbf{Ablation Study.} We demonstrate that each of our proposed components is essential for mitigating artifacts and inconsistencies, resulting in smooth consistent renderings.}
  \vspace{-5mm}
  \label{fig:ablation}
\end{figure}

\begin{wraptable}{r}{0.16\textwidth}
\centering
\vspace{-4.8mm}
\makebox[\linewidth][c]{%
\hspace{-0.045\textwidth}
\begin{minipage}{0.19\textwidth}
\caption{ \textbf{Comparison of computation efficiency}.}
\vspace{-2.5mm}
\resizebox{1\textwidth}{!}{
\begin{tabular}{l@{\hspace{0mm}}c@{\hspace{0.0mm}}}
\toprule
Modules        & {time (hr)} $\downarrow$\\ \midrule
scene init & 0.2 \\
view warping  & 0.2 \\ 
inpainting & 1.0 \\ 
4D rendering  & 0.8 \\ \midrule
\textbf{PaintScene4D}       & \textbf{2.2} \\ 
4D-fy~\cite{bahmani20244d} & 23 \\
Dream4D~\cite{zheng2024unified} & 4.5 \\
4Real~\cite{yu20244real} & 3.5 \\
VividDream~\cite{lee2024vividdream} & 3.5 \\
\bottomrule
\end{tabular}
}\label{tab:computation}
\end{minipage}%
}
\vspace{-5mm}
\end{wraptable}
\vspace{1mm}
\mypar{Runtime Analysis.} Table~\ref{tab:computation} presents a comparison of the time required to render a 4D scene from a given text prompt. Compared to object- and scene-level models, our method achieves the fastest training and rendering speed attributed to our modular and training-free framework. Additionally, to ensure a fair comparison, we normalize the rendering time based on a similar output resolution (\textit{720p}).

\subsection{Explicit Camera Control}
To assess camera control, we compare our framework with other text-to-video (T2V) models, as illustrated in Figure~\ref{fig:exp2-main}. We input the same text prompts into the T2V model twice, adjusting only the camera movement description to direct it to ``tilt towards the right'' in one case and ``move upwards'' in the other. Our observations reveal two key limitations of the T2V models. First, even with a fixed seed, the T2V model generates different scenes for each altered prompt. Second, although the model simulates an upward camera movement in the second case, it lacks explicit control over the degree of camera motion. In contrast, our approach enables explicit, consistent control over camera trajectory within the same scene and motion dynamics, leveraging 4D modeling for precise camera manipulation.

\subsection{Results on Real-World Videos}

Our proposed method exhibits strong generalization capabilities when applied to real-world video data, as demonstrated in Figure~\ref{fig:real-world}. In contrast to approaches that are strictly limited by the output of text-to-video models, our method effectively processes real video input, making it more versatile and applicable to a wider range of scenarios. A key strength of our approach is its ability to extend beyond the frames of the initial video to construct a complete 4D scene. This means that it does not just reconstruct what is visible in the given video but infers and generates additional spatial-temporal information to create a more comprehensive and dynamic scene. This ability is crucial for applications requiring realistic and immersive 4D reconstruction.


\section{Ablation Study}
\label{sec:ablation}

\begin{wraptable}{r}{0.195\textwidth}
\centering
\vspace{-4.3mm}
\makebox[\linewidth][c]{%
\hspace{-0.045\textwidth}
\begin{minipage}{0.225\textwidth}
\caption{ \textbf{Ablation Study}.}
\vspace{-2.5mm}
\resizebox{1\textwidth}{!}{
\begin{tabular}{l@{\hspace{1.5mm}}c@{\hspace{0.0mm}}}
\toprule
Model        & {CLIP $\uparrow$} \\ \midrule
w/o CIM Module  & 30.8 \\
w/o Farthest View  & 31.2 \\ 
w/o Depth Alignment & 33.9 \\ \midrule 
\textbf{PaintScene4D}       & \textbf{36.0} \\ 
\bottomrule
\end{tabular}
}\label{tab:ablation}
\end{minipage}%
}
\vspace{-5mm}
\end{wraptable}
Integrating video generation and inpainting modules is non-trivial and requires careful technical design to ensure high-quality results. Our ablation study (Figure~\ref{fig:ablation} and Table~\ref{tab:ablation}) demonstrates that the removal of key components introduces significant artifacts and inconsistencies.


\vspace{1mm}
\mypar{Depth Alignment Module.} The inclusion of the depth alignment module is crucial for maintaining the geometric consistency of the foreground. During the warping process, all frames are utilized, and any depth inconsistencies across frames result in error accumulation, leading to noticeable artifacts, particularly at the foreground boundaries.

\vspace{1mm}
\mypar{Farthest View Sampling.} In \ourwork, we select the farthest view at each step of the warping process to maximize the inpainted area. Omitting this step causes severe degradation near the edges of the foreground, such as the panda's boundary, where needle-shaped artifacts emerge due to the Gaussian splatting process.

\vspace{1mm}
\mypar{Consistent Inpainting Module.} Temporal consistency in inpainting is essential for coherent 4D scene generation. Without this module, inpainting becomes inconsistent at the boundaries of objects (\eg, the panda) across different timestamps, leading to significantly degraded renderings.

\section{Conclusion}

We introduce \ourwork, a novel framework for generating photorealistic 4D scenes from a single text prompt. Our method addresses the challenges of spatial and temporal inconsistencies and enables the generation of novel views along a user-defined camera trajectory. \ourwork outperforms existing baselines in terms of visual quality, 3D consistency, motion accuracy, and generation efficiency. Comprehensive evaluations show leading results in the generation of 4D scenes, with significantly reduced computational requirements and enhanced camera control options.

\label{sec:conclusions}

{
    \small
    \bibliographystyle{ieeenat_fullname}
    \bibliography{main}
}

\clearpage
\setcounter{page}{1}
\maketitlesupplementary
\renewcommand{\thesection}{\Alph{section}}
\renewcommand{\thefigure}{\Alph{figure}}
\renewcommand{\thetable}{\Alph{table}}
\setcounter{section}{0}
\setcounter{figure}{0}
\setcounter{table}{0}

\section{Demo Video}
\label{sec:supp:demo}

We have provided a project webpage at \href{https://paintscene4d.github.io/}{https://paintscene4d.github.io/}, including a video demonstration. This video highlights the versatility and robustness of our framework with a diverse gallery of results generated from various text prompts and scene configurations. Furthermore, to substantiate the effectiveness of our approach, the video includes detailed comparisons with baseline methods such as 4D-fy~\cite{bahmani20244d} and Dream-in-4D~\cite{zheng2024unified}. Although both 4D-fy and Dream-in-4D are designed for object-level generation, they require significantly more computational time than our approach. Furthermore, these methods are limited to generating object-level renderings, whereas our framework can render \textbf{complete scene-level generations}. These comparisons visually demonstrate the superior fidelity, consistency, and dynamicity achieved by our method across a wide range of scenarios. 

In the video, we also showcase \textbf{explicit control over camera movements} in a rendered scene, offering a significant advantage over text-to-video (T2V) models. T2V models \textit{lack direct camera control}, instead relying on implicit instructions through text prompts, which often yield inconsistent and less effective results. Additionally, T2V models generate a different scene for each iteration, even with identical prompts, limiting reproducibility. In contrast, our method supports precise manipulation of the camera trajectory within the same scene, ensuring consistency and offering greater flexibility for tailored visual outputs.

\section{More Qualitative Results}

\begin{figure*}[!ht]
  \includegraphics[width=\textwidth]{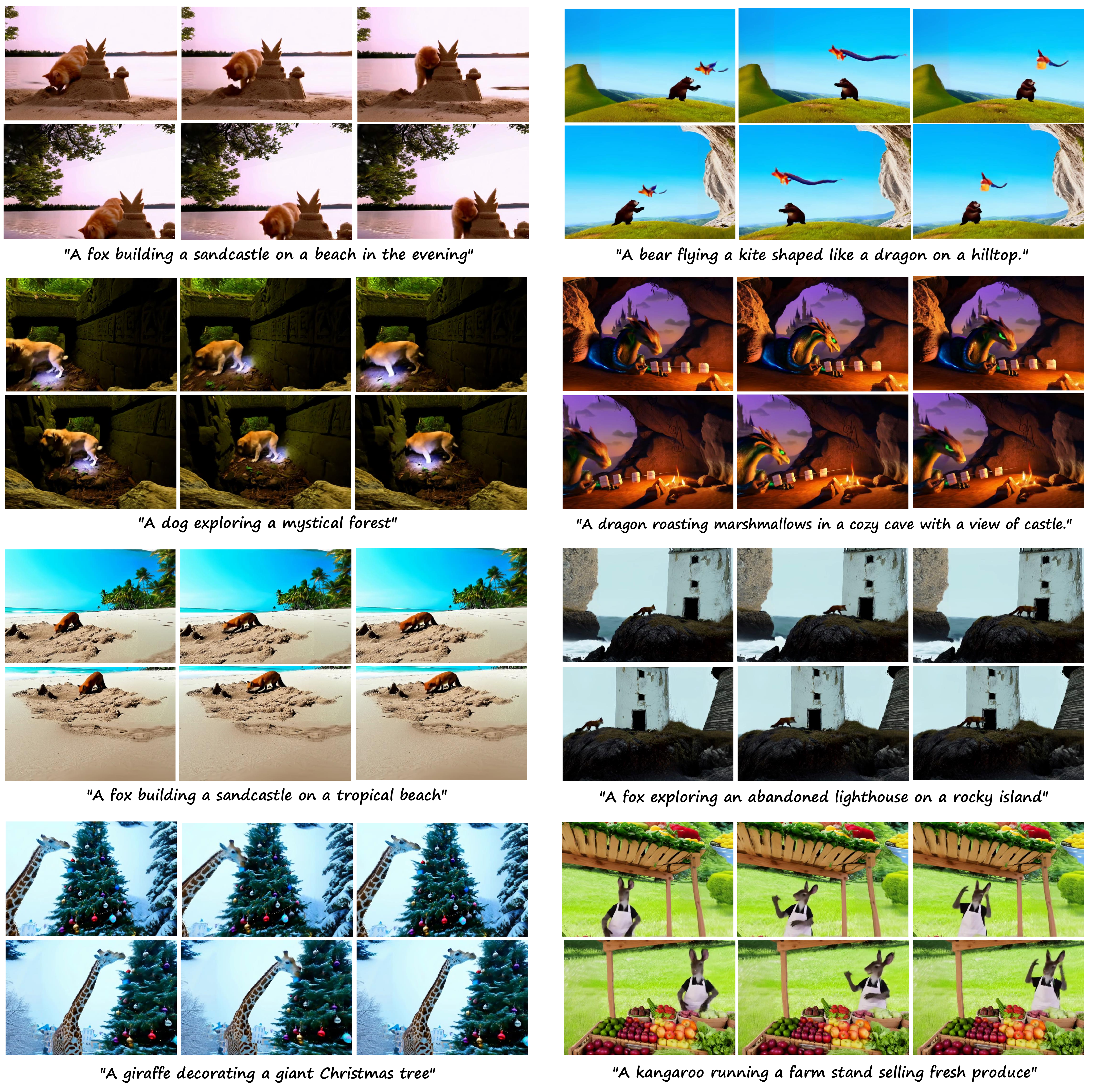}
  \caption{\textbf{Gallery of Results: } We present qualitative results of our text-to-4D generation framework, showcasing superior visual fidelity, consistent multi-view reconstructions, plausible scene compositions, and realistic dynamic motions. The horizontal axis represents the time axis and the vertical axis represents different viewpoints. A comprehensive collection of video demonstrations is provided in the supplementary materials.}
  \label{fig:supp-gallery1}
\end{figure*}

\begin{figure*}[!ht]
  \includegraphics[width=\textwidth]{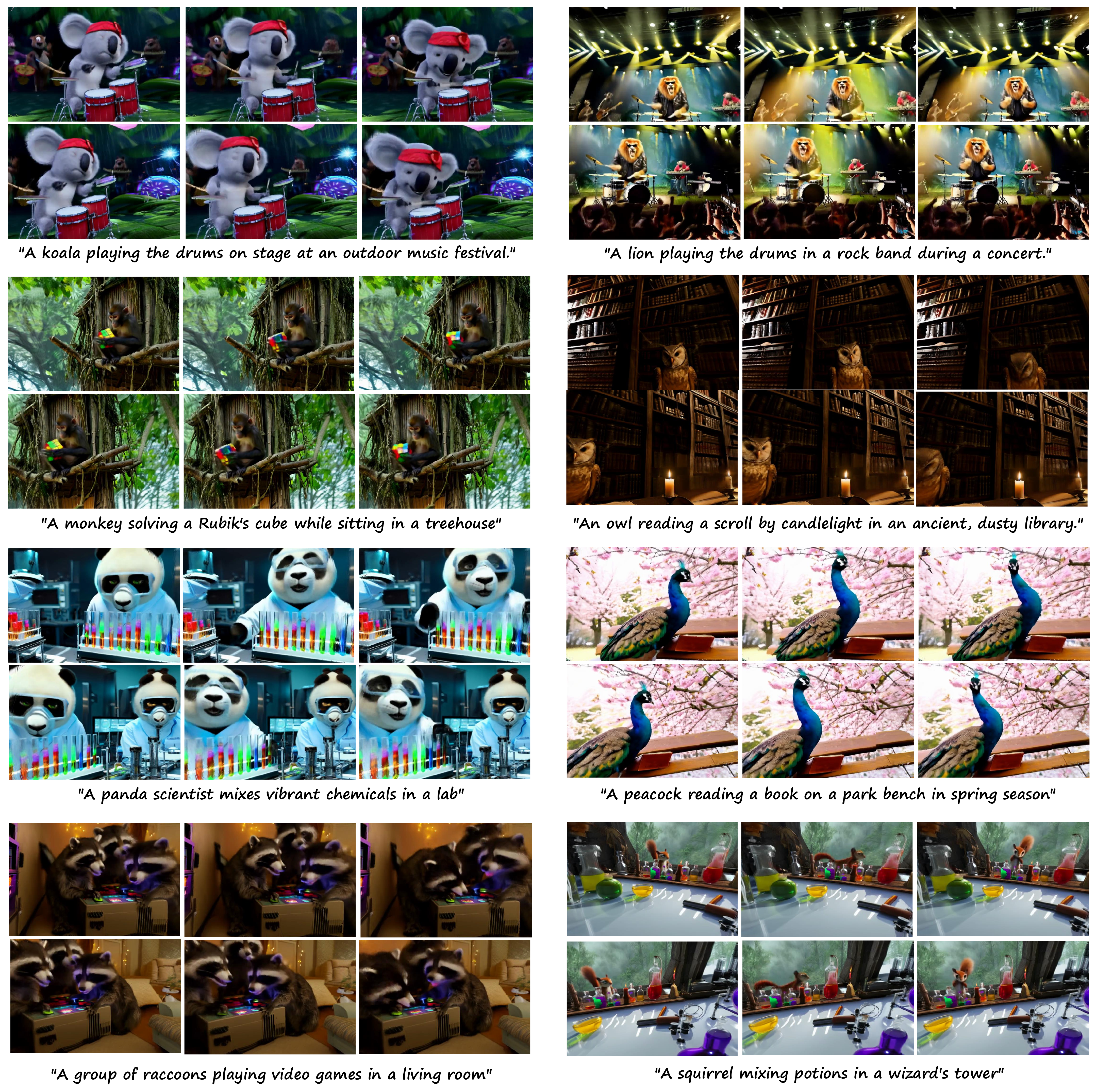}
  \caption{\textbf{Gallery of More Results. } A comprehensive collection of video demonstrations is provided in the supplementary materials.}
  \label{fig:supp-gallery2}
\end{figure*}

In Figure~\ref{fig:supp-gallery1} and Figure~\ref{fig:supp-gallery2}, we provide more examples generated using our proposed framework to demonstrate the robustness of our methodology. The horizontal axis represents the time axis, and the vertical axis represents different viewpoints. To fully appreciate the quality and diversity of our text-to-4D generation results, we strongly recommend viewing the accompanying video.

\section{\ourwork: Implementation Details} \label{sec:supp:implementation_details}

Our optimization framework comprises two stages: initially reconstructing a network of cameras, each associated with its respective view of the time frame, followed by training a 4D renderer. Specifically, we construct a network of 25 cameras and utilize videos that span 50 timestamps. All experiments are performed on a single A100 GPU. The complete warping and inpainting process, which is performed without additional training, requires approximately two hours. Following this, the 4D renderer is trained in about one hour, resulting in a total of approximately 3 hours to complete the training and generate novel views along any desired trajectory. This duration is significantly shorter than the training time required by recent state-of-the-art methods: Dream-in-4D takes over four hours, while 4Dfy takes over 20 hours, despite producing only object-level 4D renderings. To initialize the scene and establish motion priors for 4D reconstruction, we use CogVideoX-5b~\cite{yang2024cogvideox}. For depth estimation, DepthCrafter~\cite{hu2024depthcrafter} is used, as it produces consistent depth estimates across video frames, enabling reliable warping. Perspective Fields~\cite{jin2023perspective} is used to estimate the camera intrinsics for the generated video. For the segmentation model to distinguish foreground and background, we use GroundingSAM-2~\cite{ren2024grounded}.

\subsection{Hyperparameters}

\mypar{Warping and Inpainting Module.} Table~\ref{tab:supp:hyper1} demonstrates the parameters used in the warping and inpainting module. The values are carefully selected to balance efficiency and quality. The \textit{Number of Cameras} determines the multi-view coverage necessary for generating high-quality reconstructions. The \textit{Relative Depth Estimator} and \textit{Absolute Depth Estimator} are key for warping operation, with DepthCrafter used for relative depth estimation and Metric3D v2 for absolute scaling. We inpaint the missing regions multiple times and pick the best one using a CLIP~\cite{radford2021learning} based selector. The \textit{Inpainting Iterations} represents the number of times we inpaint the missing region. The \textit{Filter Size for Bilateral Filtering} sharpens the edges of a depth map necessitating better inpainting quality.

\begin{table}[!t]
\centering
  \begin{tabular}{l|c}
    \toprule
    Parameters & Value \\
    \midrule
    Number of Cameras & 25 \\
    Relative Depth Estimator & DepthCrafter~\cite{hu2024depthcrafter} \\
    Absolute Depth Estimator & Metric3D v2~\cite{hu2024metric3d}
     \\
    Inference Steps for Inpainting & 50 \\
    Inpainting Iterations & 10 \\
    Filter Size for Bilateral Filtering & [3, 5] \\
  \bottomrule
\end{tabular}
\vspace{-2mm}
\caption{Hyperparameters for the warping and inpainting module.}
\label{tab:supp:hyper1}
\vspace{-2mm}
\end{table}

\begin{table}[!t]
\centering
  \begin{tabular}{l|c}
    \toprule
    Parameters & Value \\
    \midrule
    Batch Size & 4 \\
    Number of Iterations (\textit{Coarse Training}) & 3000 \\
    Number of Iterations (\textit{Fine Training}) & 15000 \\
    Densification Until Iteration & 10000 \\
  \bottomrule
\end{tabular}
\vspace{-2mm}
\caption{Hyperparameters for 4D-GS rendering process.}
\label{tab:supp:hyper2}
\vspace{-4mm}
\end{table}

\vspace{1mm}\mypar{4D Gaussian Splatting Module.} The hyperparameters for training the 4D gaussian splatting~\cite{wu20244d} framework are presented in Table~\ref{tab:supp:hyper2}. The \textit{Number of Iterations (Coarse Training: 3000, Fine Training: 15000)} ensures robust initialization and detailed refinement. \textit{Densification Until Iteration} specifies when Gaussian points should be densely packed to model finer scene details.

\begin{table}[!t]
\centering
  \begin{tabular}{l|c}
    \toprule
    Parameters & Value \\
    \midrule
    Resolution & 720$\times$480 \\
    Number of Timestamps & 49 \\
    Number of Inference Steps & 50 \\
    Guidance Scale & 6 \\
  \bottomrule
\end{tabular}
\vspace{-2mm}
\caption{Hyperparameters for CogVideoX~\cite{yang2024cogvideox} video generation.}
\label{tab:supp:hyper3}
\vspace{-3mm}
\end{table}

\vspace{1mm}\mypar{Text-to-Video Generation.} Table~\ref{tab:supp:hyper3} presents hyperparameters for video generation with the CogVideoX text-to-video model~\cite{yang2024cogvideox}. The hyperparameters are chosen to optimize the synthesis quality and temporal consistency. The \textit{Number of Timestamps} defines the temporal resolution of the scene. The \textit{Number of Inference Steps} impacts generation fidelity.



\section{More Comparison with Other Methods}
\begin{figure}[!t]
\vspace{-2mm}
  \includegraphics[width=\linewidth]{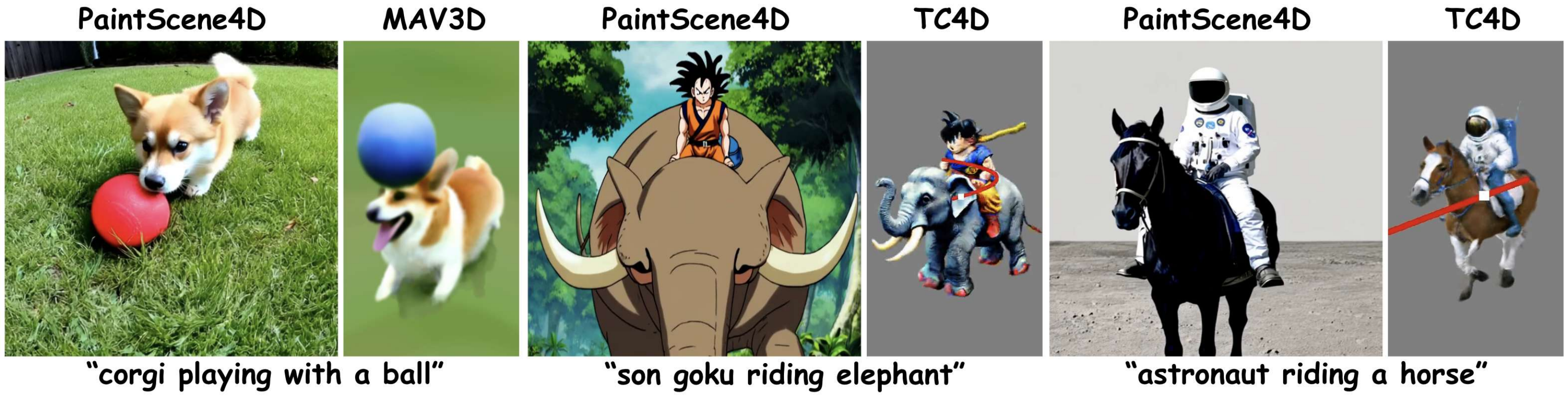}
  \vspace{-6.5mm}
  \caption{\textbf{Comparison with additional text-to-4D methods}.}
  \label{fig:tc4d}
\end{figure}

Additionally, we compare our method with other object-level approaches, such as MAV3D and TC4D, as shown in Figure~\ref{fig:tc4d}. These methods are limited to generating renderings at the object level, restricting their ability to capture broader scene contexts. Our results demonstrate that, in both cases, our method produces renderings with higher realism while also achieving faster generation times. This efficiency is notable given that MAV3D and TC4D are constrained to object-level synthesis, whereas our approach extends beyond these limitations.

\section{Discussion on Success Rate}
Our method demonstrates robustness to minor camera movements in the input video, ensuring stability in most cases. However, when there are significant deviations from a static camera setup, failure cases may occur, as illustrated in Supplementary Figure~\ref{fig:failure}. To maintain a static camera view, an appropriate prompt design can be utilized, as discussed in Section 4.1 of the main paper. With a success rate exceeding 90\%, we present our results without selective curation, further emphasizing the reliability and consistency of our approach in generating high-quality 4D scenes.

\vspace{-1.5mm}

\begin{figure}[!t]
  \includegraphics[width=\linewidth]{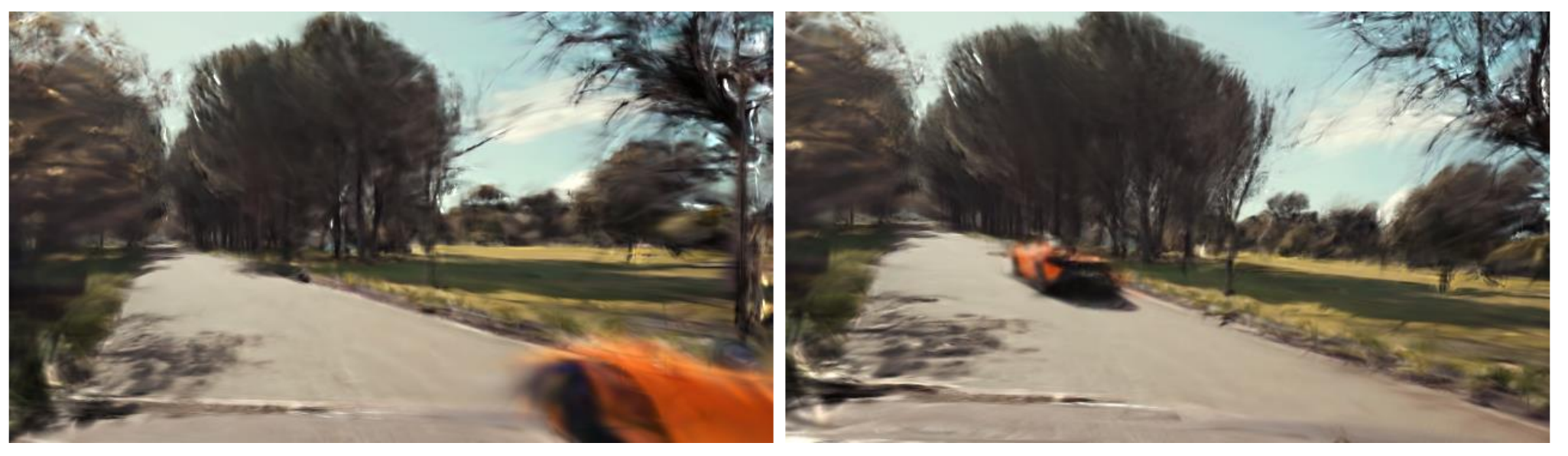}
  \vspace{-7mm}
  \caption{\textbf{Failure Case:} Our method is dependent on the assumption that the initial video generation exhibits no large camera movement. Large camera motion in the video, introduces distortions and artifacts during the subsequent rendering process, significantly affecting the visual fidelity of the final output.}
  \label{fig:failure}
  \vspace{-3mm}
\end{figure}

\section{Limitations and Future Work}\label{sec:supp:limitations}

While our method successfully generates photorealistic 4D scenes from a single text prompt, several limitations persist:  

\begin{enumerate}  

    \item \textbf{Assumption of a Static Camera:} Our approach assumes that the input video is captured from a nearly static, non-moving camera. This assumption does not always hold when using text-to-video (T2V) models, which typically offer limited control over camera dynamics. Videos with a large camera movement result in a degraded visual quality. We demonstrate the failure case in Figure~\ref{fig:failure}. Therefore, extending our framework to accommodate videos with more camera movements represents a promising direction for future work.  

    \item \textbf{Lack of Explicit 3D Foreground Modeling:} Our current method does not explicitly model the 3D structure of the foreground. Instead, we rely on an inpainting model to fill in gaps at the boundaries of the foreground, which means that the model does not possess a comprehensive understanding of the 3D geometry of the scene. A more advanced approach could involve explicitly separating the foreground from the background and modeling the 3D structure of the foreground, potentially using methods like SV4D~\cite{xie2024sv4d}.  

    \item \textbf{Challenges with Rapid Motion:} Our approach struggles to handle rapid movements in the video due to the limitations of current 4D rendering techniques. Advancements in this area would likely enhance the rendering quality of our method and enable better handling of fast motion.  

    \item \textbf{Segmentation Errors and Artifacts:} If the segmentation model fails to accurately distinguish the character or foreground from the background, it can introduce significant errors during the warping and inpainting processes. These inaccuracies accumulate over successive stages, leading to noticeable artifacts in the final rendered output. One common issue is the presence of double geometry, where duplicated or misaligned structures appear in the scene, reducing the overall visual quality and realism of the generated 4D representation. 

\end{enumerate}  

\end{document}